\documentclass[11pt]{article}

\usepackage{balance} 
\usepackage{soul}
\usepackage{authblk}
\usepackage{booktabs}
\usepackage{multirow}
\usepackage{graphicx}
\usepackage[numbers]{natbib}
\usepackage{authblk}   
\usepackage{hyperref}  
\usepackage{footmisc}  
\usepackage[english]{babel}
\usepackage{amsthm}

\theoremstyle{definition}
\newtheorem{definition}{Definition}[section]

\date{}

\title{Birds of a Different Feather Flock Together: \\ Exploring Opportunities and Challenges in \\ Animal-Human-Machine Teaming}

\author[1,2]{Myke C. Cohen}
\author[3]{David A. Grimm}
\author[4]{Reuth Mirsky}
\author[1]{Xiaoyun Yin}

\affil[1]{Arizona State University, Mesa, AZ}
\affil[2]{Aptima, Inc., Woburn, MA}
\affil[3]{Georgia Institute of Technology, Atlanta, GA}
\affil[4]{Tufts University, Medford, MA}

\begin{document}

\maketitle

\begin{abstract}
Animal-Human-Machine (AHM) teams are a type of hybrid intelligence system wherein interactions between a human, AI-enabled machine, and animal members can result in unique capabilities greater than the sum of their parts. This paper calls for a systematic approach to studying the design of AHM team structures to optimize performance and overcome limitations in various applied settings. We consider the challenges and opportunities in investigating the synergistic potential of AHM team members by introducing a set of dimensions of AHM team functioning to effectively utilize each member's strengths while compensating for individual weaknesses. Using three representative examples of such teams---security screening, search-and-rescue, and guide dogs---the paper illustrates how AHM teams can tackle complex tasks. We conclude with open research directions that this multidimensional approach presents for studying hybrid human-AI systems beyond AHM teams. 
\end{abstract}


\section{Introduction}

Consider a Blind or Visually Impaired (BVI) person training to be assisted by a guide dog. When the pair reaches an obstacle along their path, they can aim to employ a learned protocol: the dog stops, then the BVI person investigates the cause for the stop, be it a wall, a stair, etc. With modern vision technologies, the pair could be equipped with a designated camera and a vision algorithm that automatically identifies the reason for the dog’s stop \cite{ichikawa2022voice}, reducing the need for the BVI person to investigate manually. 
Such a system can be considered as an animal-human-machine (AHM) team that functions as a synergistic unit that can tackle complex tasks and challenges. Conversely, the absence of any of the members makes AHM teams less capable: without a vision system, a BVI person must exert inefficient and potentially risky efforts to interpret why their guide dog pair made a stop; even the most cutting-edge robots cannot fully replace a guide dog \cite{morris2003robotic,sakhardande2012smart,mirsky2021seeing, hwang2024towards}; and obviously, the team's goal ceases to exist without the BVI person. 

\begin{figure}
    \centering
    \includegraphics[width=0.75\linewidth]{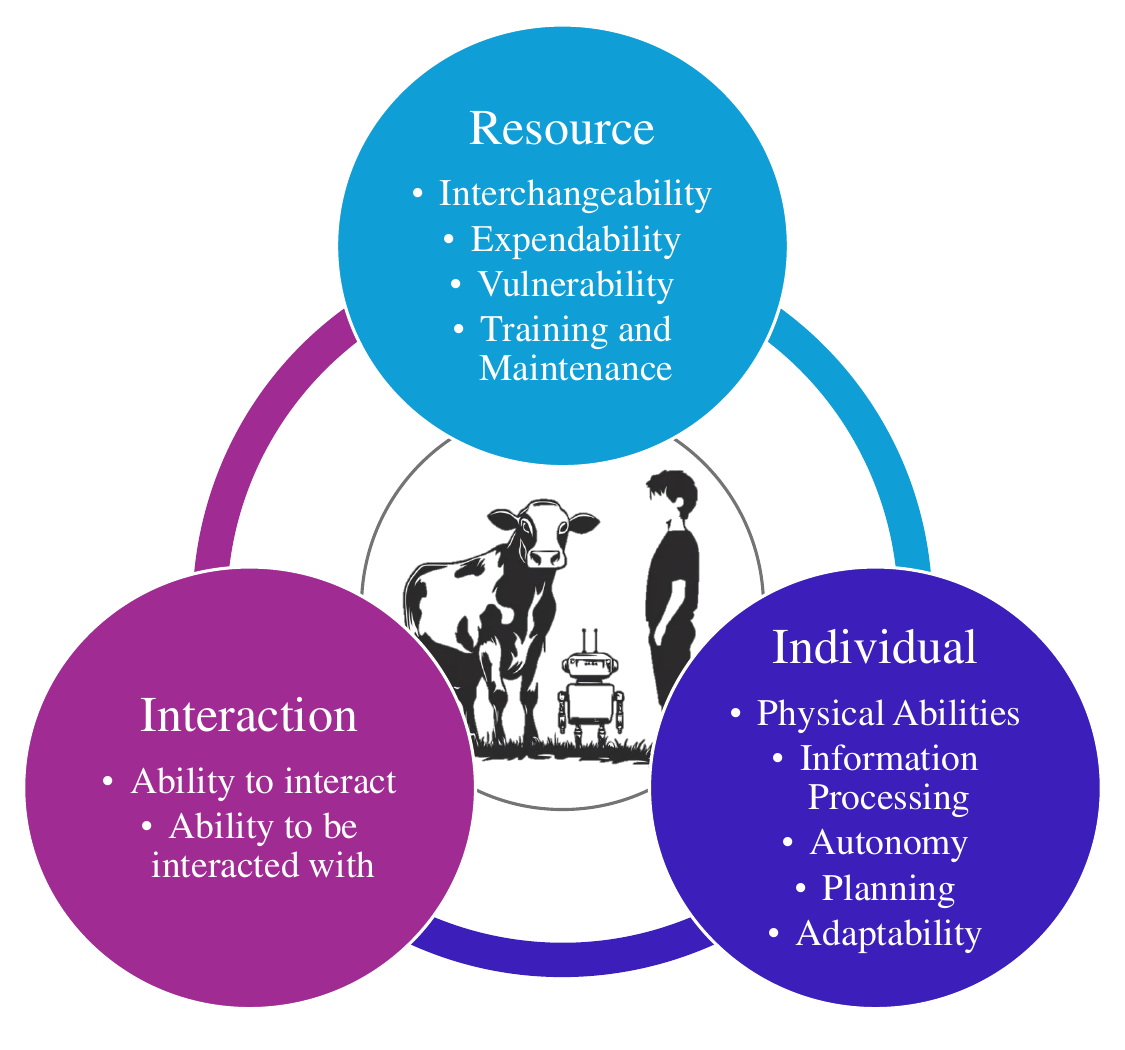}
    \caption{The properties of Animal-Human-Machine team.}
    \label{fig:ahm}
\end{figure}

In the ideal scenario, each member of an AHM team is given a role that maximizes their respective strengths and compensates for individual weaknesses to optimize the team's functionality. However, in high-stakes, rapidly changing scenarios, individual workloads are often exceeded, whether through anticipated or unanticipated circumstances. In those situations, flexible reallocation is needed to redistribute responsibilities across other team members \cite{bye1999human}. 
This paper investigates AHM teams as a type of a multiagent system (MAS) and discusses the unique research challenges and opportunities in this setup. 
We define an AHT team as follows:
\begin{definition}
\label{def:ahm}
An Animal-Human-Machine team is a multiagent system with three or more members: at least one is an animal, one is a human, and one is a machine.
Each member has agency grounded in unique sensing, decision-making, and acting capabilities.
\end{definition}

 Previous work has also highlighted some conceptual similarities between domestic animals and machines \cite{darling2021new}. As some robots are of organic matter  \cite{blackiston2023biological} and animals are incorporated into machines \cite{sanchez2015locomotion}, we might reach a point where the distinction between the different agents is blurred \cite{mazis2008humans}. At this time, however, we assume that the boundaries between humans, animals, and machines are clear. 
 We note that this paper focuses on the technological aspects of AHM rather than its legal or ethical implications, which are not negligible. Interested researchers can refer to \cite{mazis2008humans,schaerer2009robots,hou2024power}.

We move on to discuss AHM from the lens of related work, including team functional allocation from industrial/organizational psychology and computational work linking human-animal and human-agent interaction paradigms. 


\section{Related Works}



In organizational psychology, teams are often defined as groups of people with a shared sense of identity who collaborate with distinct roles for a limited time to achieve a common goal \cite{salasUnderstandingTeamPerformance1992}. By this definition, AHM teams seem paradoxical due to having non-human members. But treating components like sense of identity or role distinction as variable elements of teaming (i.e., ``teamness''; \cite{cookeTeamsTeamnessFuture2024}) can aid in understanding a broad range of collaborations. Consider ad-hoc teamwork, where amorphous team compositions may result in a lack of clear shared identities among collaborators
\cite{stone2010ad}. Treating shared identity as a variable has borne algorithmic solutions that aid teammates in establishing common ground, inferring each other's capabilities, and interpreting other \textit{agents}' underlying cooperative intents \cite{grosz1999evolution,mirsky2022survey}. Similarly, considering  ``humanness'' as a spectrum rather than a requirement of teaming has benefited the recent surge in research and development of human-machine systems wherein people and machines collectively achieve feats they could not apart \cite{oneillHumanAutonomyTeaming2022, cookeTeamsTeamnessFuture2024}. The same logic has long been used to describe human-animal systems \cite{harawayCompanionSpeciesManifesto2003}. 

Human-animal teams are increasingly used as a metaphor for human-machine teams. Both are interactive systems that achieve joint outcomes despite having different cognitive processes between animal, human, and machine members \cite{lumUnderstandingHumanAutonomy2023, coeckelberghHumansAnimalsRobots2011, kruegerHumanDogRelationships2021, billings2012human}. Human-animal teams and human-machine teams also share intractable physical and cognitive asymmetries between each member type, limiting the tasks they can perform and their understanding of the team's overarching goals \cite{phillipsHumananimalTeamsAnalog2016}. These asymmetries are amplified in AHM teams due to the concurrent presence of at least two types of non-human teammates \cite{gerencserIntegrationDogsCollaborative2016},
defining a complex range of research paradigms and application domains. For example, how to support the development of shared mental models between human and machine teammates \cite{baker2011bayesian, fioreTransdisciplinaryTeamScience2023, schelbleLetsThinkTogether2022, andrewsRoleSharedMental2023, narayananInfluenceHumanAITeam2024}; however, the impacts of these efforts will likely be limited in full AHMs due to animal cognitive limitations. Conversely, the evolutionary origin of human-animal relationships results in interaction dynamics that are more social and emotional than human-robot ones \cite{coeckelberghHumansAnimalsRobots2011}. 


One quality that AHM teams ostensibly share is role heterogeneity, i.e., each member is assigned a unique set of tasks to fulfill a role within a team \cite{belbinTeamRolesWork1993}. Role heterogeneity often results from delegating tasks to individuals based on their abilities. In all-human teams like surgical teams, task assignment decisions are based on limited pools of team members who can fulfill roles like surgeon, anesthesiologist, or nurse at a given time \cite{carayonWorkSystemDesign2006, gorman2020measuring}. However, further training can enable today's nurse to be assigned the role of surgeon in the future. In contrast, such decisions in AHM teams are limited by ethical, legal, and technological concerns that come with assigning non-human members to potentially consequential tasks \cite{bonfantiSnifferDogsEmerging2014, coman2018ai, parasuramanHumansStillVital2008}. 

In human-machine systems, Fitts' list \cite{fittsHumanEngineeringEffective1951, de2014fitts} is an influential framework that operates on the principle of non-overlapping task allocations based on the relative strengths and weaknesses of people and machines (often called ``HABA-MABA'', short for ``humans-are-better-at/machines-are-better-at''). Current functional allocation paradigms in human teams and human-machine teams are still predominantly based on individual members' capabilities \cite{howardFrameworkHumanAlgorithmTeaming2023,caldwellAdvancesHumanAutomationCollaboration2019, laiScienceHumanAIDecision2021, crawfordConfiguralTheoryTeam2013}. This premise remains true amid calls for more nuanced consideration of interactions between complex human and machine teammates rather than individual capabilities alone  \cite{caldwellAdvancesHumanAutomationCollaboration2019, johnsonCooperativeRoboticsCentral2011, feighRequirementsEffectiveFunction2014, metcalfeSystemicOversimplificationLimits2021}.

To illustrate, in an AHM security screening setting, trained sniffing dogs are fairly reliable in detecting scent markers of various illicit items \cite{jezierskiEfficacyDrugDetection2014, myersDoghandlerTeamDetection1992}; ``sniffing'' capabilities for machines have been developed and deployed as machine teammates or alternatives to dogs, and can be more or less effective depending on the target items \cite{furtonComparingOlfactoryCapabilities2022}; last, and the least useful candidate for the scent detection task, is the human. However, for other tasks like physical inspection of suspicious items, people may be better than machines, who in turn are likely better than dogs. The original Fitts' list does not account for the strengths and weaknesses of animals; nonetheless, its strengths-and-weaknesses framework correctly predicts current AHM security screening configurations \cite{furtonComparingOlfactoryCapabilities2022}. Other AHM teaming contexts, like field search-and-rescue and enhanced guide dogs, have less structured team goals and tasks. Thus, additional factors are needed to inform functional allocation decisions. 


\section{Functional Allocation Dimensions}

We propose a set of AHM team functional allocation considerations based on the potential strengths and weaknesses of AHM teams. These are divided into three categories, summarized in Figure \ref{fig:ahm}: (1) the \textit{Individual} roles and functions that AHM team members will need to fill to accomplish the team's shared goals; (2) the \textit{Interaction} capabilities required for AHM teammates to function as a unit; and (3) the \textit{Resource} considerations for using AHM team designs as opposed to more traditional teams.

\subsection{Individual Dimensions}
A primary consideration in the design of AHM teams is how the unique capabilities of humans, animals, and machines can be applied to individual tasks needed to accomplish team goals. Taskwork-related functional allocation considerations include the following:

\textbf{Physical Requirements.} Physical requirements underpin functional allocation setups in many AHM systems, with several key trade-offs. For example, high-exertion, low-precision tasks can be primarily assigned to working animals or simple machines (e.g., animal-powered plowing). High-precision tasks can be assigned to humans or specialized machines, depending on the repetitiveness and intensity of the work.

\textbf{Information Processing Requirements.} Modern AHM functional allocation tends to be driven by the unique capabilities of each member to fulfill information processing requirements. Animals in such teams can often be trained to respond to scents (e.g., sniffing dogs) or sounds (e.g., navy dolphins) that they are evolutionarily well-adapted to detect. Machine sensors (e.g., magnetometers and thermal cameras) can detect information streams beyond human or animal capabilities, and algorithms can analyze specific information at astounding rates and volumes. Humans can cover a wide range of information-processing tasks and are often expected to handle those involving sociocultural contexts. 
    
\textbf{Level of Autonomy.} Differences in AHM agent abilities to independently fulfill task roles often restrict possible functional allocation configurations and require careful consideration \cite{coman2018ai, liu2018learning}. For example, AI agents may prove to be better teammates and engage more proactively by assigning them with higher agency, like interrupting other agents \cite{mannem2023exploring, monk2008effect, mckenna2020sorry, chiang2014personalizing, horvitz2005balancing}. 

\textbf{Planning Capabilities.} For teams to function, each member must have a sufficient understanding of their tasks, the steps needed to complete them, and how to complete these tasks and steps. Though human agent(s) would likely be in charge of most planning and assigning tasks within an AHM team, it is still necessary for other non-human teammates to understand their given task to accomplish it. Moreover, all agents may benefit from a better understanding of their teammates' beliefs, desires, and intentions \cite{georgeff1999belief, shvo2020epistemic}. Several animals demonstrate Theory of Mind capabilities \cite{krupenye2019theory}, and relevant AI research can boost machine agents' ability for shared planning \cite{grosz1999evolution, d2004dmars, kamar2009incorporating}. Because it is widely recognized as a necessary aspect of team effectiveness \cite{burke2006understanding, marks2001temporally, driskell2018foundations}, planning capabilities are also closely tied to interaction dimensions covered in the following section.

\textbf{Adaptability.} Finally, in dynamic and high-risk work domains, such as search and rescue, AHM team members' abilities to learn and adapt taskwork processes when needed are another important taskwork consideration. This is related to the agent's training requirement (a resource dimension in Section \ref{sec:resouce}) but focuses less on the external costs of training or equipping a teammate for adaptability and more on the agent's ability to re-plan in real-time \cite{knott2021evaluating, newman2024bootstrapping}.




\subsection{Interaction Dimensions}
Interaction dimensions comprised two high-level dimensions: the ability to actively interact with other AHM team members, and the ability to respond to team members' interactions with them. Unlike individual taskwork dimensions, in which teammate characteristics may be measured at the level of an individual teammate, these interaction dimensions rely on more than one teammate to be measured. The team competencies described in this section may be considered to be generalizable team skills, such as communication, coordination \cite{salas2008communicating, tannenbaum2020teams, semeraro2023simpler}, or trust \cite{costaTrustTeamsRelation2001}. 

\textbf{Ability to Interact.} 
The ability to interact with other members of an AHM team involves communicating with teammates to coordinate individual tasks \cite{cookeInteractiveTeamCognition2013}. On the one hand, communication can occur between human controllers and animals, where the human can provide overarching goals for animal agents; similarly, control machines (e.g., drones) may be operated remotely from a command station. Communication will possess challenges since each agent will naturally communicate in different ways. This may impact how warning signs are communicated to each teammate. There are solutions that can be gleaned from existing research, wherein animals can be taught to respond to computer-mediated signals, such as shock signals \cite{bozkurt2014toward}, when the animal deviates from safe parameters. On the other hand, the coordination aspect of effective AHM teaming lies in the timely exchange of information among team members. While coordination is a widely studied aspect of team effectiveness in team research \cite{entin1999adaptive, gorman2010team, burtscher2010managing}, it is especially important in AHM teams, where three different types of agents must coordinate their activities effectively. This is why the ability to interact with different types of teammates can require extensive resources to train or develop the necessary skills. 

Other interaction-based abilities include \textit{social intelligence}, \textit{co-learning}, and \textit{trust}. Social intelligence refers to the knowledge of other agents' emotional states and knowing when to provide information to others. It is well documented that animals (e.g., dogs) possess strong social intelligence, which allows them to understand their owners' feelings and even intentions \cite{cooper2003clever, maginnity2014visual, hare2005human}. Robots potentially develop similar capabilities, making social intelligence a pertinent characteristic of AHM teams. Learning may be unique in AHM teams because, in contrast to unidirectional human-to-animal or human-to-machine training (e.g., operant conditioning), the significant differences between teammate's learning capabilities require a more bi-directional co-learning approach. Finally, trust is the belief that teammates can assist teammates in achieving their goals in situations of high uncertainty and complexity \cite{lee2004trust, lewis2018role, khavas2020modeling}. Due to different cognitive and emotional capabilities, trust may develop differently in animals and humans than in traditional teams.

\textbf{Ability to be Interacted With.}
These components of AHM teaming describe how a teammate can be an effective teammate by facilitating effective interactions. 
First, as described above, proper trust helps enable teammates to interact with others. Being a trustworthy teammate can assist in understanding how one will react in a particular situation. This is related to the concept of \textit{reliability}, which may develop differently in machines and animals. In the former, reliability may describe how well a machine can perform a task without breaking. In the latter, it may describe how well an animal performs a task and adheres to training and the humans' command. For example, some guide dogs get more prone to pulling, so their handler is trained to expect and counter this behavior \cite{craigon2017she}.

Trust and reliability are similar to the concept \textit{scrutability} of a team member, which is the ability to thoroughly examine and understand a system \cite{kay2006scrutable}. It is expected that a trustworthy and reliable team member is one that its team members can well understand. The characteristic of \textit{transparency} may also underlie the ability to be interacted with effectively. Transparency operates at a level of both behavior and intent, as it had been defined previously within the context of AI \cite{lyons2013being}. However, the transparency of an animal can be unique from a machine's. 
Finally, \textit{predictability} is the ability to anticipate the actions of a machine or animal. By possessing a better understanding of the interaction capabilities of humans and machines, it is hoped that functions can be effectively allocated among the different members of an AHM team.

\subsection{Resource Dimensions}
\label{sec:resouce}
These dimensions cover potential difficulties, vulnerabilities, or expenditures of time and money associated with AHM teams. 

\textbf{Interchangeability.}
We define interchangeability as the ability to swap individual team members while maintaining previous levels of team functioning and effectiveness. This is potentially costly in AHM teams where different members have varying levels of stamina, movement, or fluidity capabilities and may possess specific training or pre-programmed abilities. These qualities may make interchangeability more complex in AHM than in traditional teams. The concept of interchangeability in AHM has been widely researched and aligns with the original HABA-MABA principles~\cite{fittsHumanEngineeringEffective1951}.

\textbf{Expendability.}
The next potential cost is expendability, which describes how any particular agent's value can be evaluated through a cost/benefit analysis. It is expected that AHM teams will have high expandability costs, as there is a natural tradeoff in abilities: humans, animals, and machines are inherently different. Considering that AHM teams are likely to engage in high-risk scenarios (e.g., search-and-rescue, military), expendability is likely to be a common consideration for AHM teams. One potential avenue to investigate the expandability of agents in a team is assigning them with Shapely values according to their different abilities \cite{shapley1953value}.

\textbf{Vulnerability.}
Another cost consideration is \textit{vulnerability}, defined as how susceptible each team member is to harm in an AHM task. The team must estimate each agent's vulnerability, especially in perilous tasks, such as search-and-rescue scenarios. 

\textbf{Training \& Maintenance.}
Finally, there is the \textit{cost to train and maintain} agents in an AHM team. It is common for humans and animals to take months or years to train effectively on their unique skills, so time investment will influence how animals are incorporated into an AHM team. Furthermore, human investments also require constant labor upkeep through continued education and compensation costs. Conversely, machine learning is also resource-heavy and often requires specialized hardware to be effective. Given current technology, some tasks might be learnable by machines in theory but would require too much training data in practice.
\section{Use-Cases}
We consider some real-life examples of AHM teaming in light of the dimensions of our proposed functional allocation framework, namely security screening, search and rescue, and machine-enhanced guide dog setups. 

\subsection{Security Screening}
In security screening systems, the goal is to detect and address the presence of potentially dangerous items. These often involve two screening steps, i.e., preliminary screenings and manual inspections. 
\textit{Information processing capabilities} plays a crucial role: Sniffing dogs can only serve as preliminary screening agents for detecting suspicious odors. Machines also perform only preliminary screening tasks with specialized sensors like metal detectors and X-ray scanners. 
Interaction considerations help explain the presence of several human roles in AHM security screening teams, which can be differentiated depending on which non-human agents they interact with: (1) dog handlers, who respond to sniffing dog reactions to screened items and manage sniffing dogs \cite{myersDoghandlerTeamDetection1992}; (2) x-ray operators, who interact with x-ray screening machines, manipulating and interpreting images to flag suspicious items or persons for further inspection \cite{hattenschwilerDetectingBombsXRay2019}; and (3) manual inspectors, who manually inspect flagged items. 
%
\textit{Training and maintenance costs} are intertwined with limited role interchangeabilities. For example, dog handlers often only handle specific sniffing dogs to accommodate \textit{reliability} and \textit{trust} considerations. In contrast, x-ray operators typically rotate between machines they are trained to operate, having general performance expectations from the technology in general. Sniffing dogs themselves must be extensively trained, with a cost compounded by the need to train them with dog handlers as a human-animal dyad. Because security screening tasks normally occur in places like airports where there is a high volume of screening tasks in a fast-paced environment, managing dog exhaustion is also a concern. However, although ``sniffing machines'' may be an alternative, the deployment and maintenance costs of these nascent technologies remain an obstacle today. 

\subsection{Search and Rescue}

In search and rescue missions, the synergy between animals and humans is crucial for maximizing success in life-threatening situations. 
For instance, the ability of both animals and humans to quickly \textit{process} and \textit{adapt} to changing environments is essential when navigating hazardous disaster areas like those seen in the aftermath of Hurricane Katrina (Kassraie, 2021).

Effective \textit{communication} and mutual \textit{trust} between animal and human team members form the backbone of successful operations. Clear, consistent signals and commands enable seamless coordination, while trust allows team members to rely on each other's unique capabilities. This is particularly important when leveraging complementary skills, such as a rescue dog's superior sense of smell combined with a human handler's problem-solving abilities, to locate victims in complex rubble structures.

The shared capacity for information processing among animals and humans is a critical factor in search and rescue effectiveness. Both must process the environment and develop accurate spatial representations to navigate treacherous areas efficiently. This cognitive skill, coupled with stress resilience and interspecies empathy, allows teams to maintain high performance under extreme pressure. By focusing on these attributes in training and team composition, search and rescue organizations can significantly enhance their ability to locate and save victims in disaster scenarios, addressing the fundamental question of how individual spatial cognition contributes to overall team performance in applied settings.

\subsection{Guide Dogs}
Consider the use-case presented at the beginning of the paper of a guide dog setup extended with AI capabilities: Modern vision systems can identify these different scenarios and alert the person regarding the reason for stopping using voice. This example highlights each individual's unique contribution to this team: the person provides high cognitive ability and decision-making; the dog provides sensing and physical responsiveness; the vision systems provide enhanced sensing and inference capabilities. \textit{Interchangeability} in this scenario is uniquely low, as the person cannot replace the role of the dog, the dog cannot speak in detail about why it stopped, and the vision system cannot physically lead a person. \textit{Expendability} is high because dog training is expensive and cannot be easily replaced. The \textit{ability to communicate} and \textit{be communicated with} are also especially important in this context, as each member's actions are highly coupled with the other teammates' actions. 
\section{Conclusion}
This paper delineates the dimensions involved in AHM teaming, planning, and execution with respect to current MAS research. It further discusses three promising use cases that help highlight the uniqueness of such teams. We encourage MAS researchers to identify how their work can be utilized in AHM teams, as it holds potential mutual benefits: improving the team's performance while providing insights to promote research further. For example, a machine's alarm can help inform a human, but at the cost of disturbing the animal. Research on the cost of communication can be utilized to improve the decision of when to sound an alarm while providing new and exciting insights from this real-world setup.

\small
\bibliography{MAIN}
\bibliographystyle{unsrtnat}

\end{document}